Pawan Kumar Singh[1], Iman Chatterjee[2], Ram Sarkar[3] and Mita Nasipuri[4]

# Handwritten Script Identification from Text Lines

**Abstract:** In a multilingual country like India where 12 different official scripts are in use, automatic identification of handwritten script facilitates many important applications such as automatic transcription of multilingual documents, searching for documents on the web/digital archives containing a particular script and for the selection of script specific Optical Character Recognition (OCR) system in a multilingual environment. In this paper, we propose a robust method towards identifying scripts from the handwritten documents at text line-level. The recognition is based upon features extracted using Chain Code Histogram (CCH) and Discrete Fourier Transform (DFT). The proposed method is experimented on 800 handwritten text lines written in seven *Indic* scripts *namely*, *Gujarati*, *Kannada*, *Malayalam*, *Oriya*, *Tamil*, *Telugu*, *Urdu* along with *Roman* script and yielded an average identification rate of 95.14% using Support Vector Machine (SVM) classifier.

**Keywords:** Script Identification, Handwritten text lines, *Indic* scripts, Chain Code Histogram, Discrete Fourier Transform, Multiple Classifiers

## 1 Introduction

One of the major Document Image Analysis research thrusts is the implementation of OCR algorithms that are able to make the alphanumeric characters present in a digitized document into a machine readable form. Examples of the applications of such research include automated word recognition, bank check processing, and address sorting in postal applications etc. Consequently, the vast majority of the OCR algorithms used in these applications are selected based upon a priori knowledge of the script and/or language of the document under analysis. This assumption requires human intervention to select the appropriate OCR algorithm, limiting the possibility of completely automating the analysis process, especially when the environment is purely multilingual. In this scenario, it is very necessary to have the script recognition module before applying such document into appropriate OCR system.

[1] Department of Computer Science and Engineering, Jadavpur University, Kolkata, India
pawansingh.ju@gmail.com
[2] Department of Computer Science and Engineering, Netaji Subhash Engineering College, Kolkata, India
imanchatterjee9@gmail.com
[3] Department of Computer Science and Engineering, Jadavpur University, Kolkata, India
raamsarkar@gmail.com
[4] Department of Computer Science and Engineering, Jadavpur University, Kolkata, India
mitanasipuri@gmail.com

In general, script identification can be achieved at any of the three levels: (a) Page-level, (b) Text-line level and (c) Word-level. In comparison to page or word-level, script recognition at the text line-level in a multi-script document may be much more challenging but it has its own advantages. To reliably identify the script type, one needs a certain amount of textual data. But identifying text words of different scripts with only a few numbers of characters may not always be feasible because at word-level, the number of characters present in a single word may not be always informative. In addition, performing script identification at word-level also requires the exact segmentation of text words which is again an exigent task. On the contrary, identifying scripts at page-level can be sometimes too convoluted and protracted. So, it would be better to perform the script identification at text line-level than its two counterparts.

A detailed state-of-the-art on *Indic* script identification described by P. K. Singh *et al.* [1] shows that most of the reported studies [2-8], accomplishing script identification at text line-level, work for printed text documents. G. D. Joshi *et al.* [2] proposed a hierarchical script classifier which uses a two-level, tree based scheme for identifying 10 printed *Indic* scripts *namely*, *Bangla*, *Devanagari*, *Gujarati*, *Gurumukhi*, *Kannada*, *Malayalam*, *Oriya*, *Tamil* and *Urdu* including *Roman* script. A total of 3 feature set such as, statistical, local, horizontal profile are extracted from the normalized energy of log-Gabor filters designed at 8 equi-spaced orientations ($0°$, $22.5°$, $45°$, $77.5°$, $90°$, $112.5°$, $135.5°$ and $180°$) and at an empirically determined optimal scale. An overall classification accuracy of 97.11% is obtained. M. C. Padma *et al.* [3] proposed to develop a model based on top and bottom profile based features to identify and separate text lines of *Telugu*, *Devnagari* and *English* scripts from a printed tri-lingual document. A set of eight features (i.e. bottom max-row, top-horizontal-line, tick-component, bottom component (extracted from the bottom-portion of the input text line), top-pipe-size, bottom-pipe-size, top-pipe-density, bottom-pipe-density) are experimentally computed and the overall accuracy of the system is found to be 99.67%. M. C. Padma *et al.* [4] also proposed a model to identify the script type of a trilingual document printed in *Kannada*, *Hindi* and *English* scripts. The distinct characteristic features of said scripts are thoroughly studied from the nature of the top and bottom profiles. A set of 4 features *namely*, profile_value (computed as the density of the pixels present at top_max_row and bottom_max_row), bottom_max_row_no (the value of the attribute bottom_max_row), coeff_profile, top_component_density (the density of the connected components at the top_max_row) are computed. Finally, *k*-NN (*k*-Nearest Neighbor) classifier is used to classify the test samples with an average recognition rate of 99.5%. R. Gopakumar *et al.* [5] described a zone-based structural feature extraction algorithm for the recognition of South-*Indic* scripts (*Kannada*, *Telugu*, *Tamil* and *Malayalam*) along with *English* and *Hindi*. A set of 9 features such as number of horizontal lines, vertical lines, right diagonals, left diagonals, normalized lengths of horizontal lines, vertical lines, right diagonals, left diagonals and normalized area of the line image are computed for each text line image. Finally, the classification accuracies of 100% and 98.3% are achieved using *k*-NN and SVM (Support Vector Machine) respectively. M. Jindal *et al.* [6] proposed a script identification approach for *Indic* scripts at text line-level based upon features extracted using Discrete Cosine Transform (DCT) and Principal Component Analysis (PCA) algorithm. The proposed method is tested on printed document images in 11 major Indian languages (*viz.*, *Bangla*, *Hindi*, *Gujarati*, *Kannada*, *Malayalam*, *Oriya*, *Punjabi*, *Tamil*, *Telugu*, *English* and *Urdu*) and 95% recognition accuracy is obtained. R. Rani *et al.* [7] presented the effectiveness of Gabor filter banks using *k*-NN, SVM and PNN (Probabilistic Neural Network) classifiers to identify the scripts at text-line level from trilingual documents printed in *Gurumukhi*, *Hindi* and *English*. The experiment shows that a set of 140 features based on Gabor filter with SVM classifier achieve the maximum recognition rate of 99.85%. I. Kaur *et al.* [8] presented a script identification work for the identification of English and Punjabi scripts at text-line level through headline and characters density features. The approach is thoroughly tested for different font size images and an average accuracy of 90.75% is achieved. On the contrary, researches made on handwritten documents are only a few in number. M. Hangarge *et al.* [9] investigated texture pattern as a tool for determining the script of handwritten document image, based on the observation that text has a distinct visual texture. A set of 13 spatial spread features of the three *Indic* scripts *namely*, *English*, *Devanagari* and *Urdu* are extracted using morphological filters and the overall accuracies of the proposed algorithm are found to be 88.67% and 99.2% for tri-script and bi-script classifications respectively using *k*-NN classifier. P. K.

Singh *et al.* [10] proposed a texture based approach for text line-level script identification of six handwritten scripts *namely*, *Bangla*, *Devanagari*, *Malayalam*, *Tamil*, *Telugu* and *Roman*. A set of 80 features based on Gray Level Co-occurrence Matrix (GLCM) is used and an overall recognition rate of 95.67% is achieved using Multi Layer Perceptron (MLP) classifier. To the best of our knowledge, script identification at text line-level considering large number of *Indic* handwritten scripts does not exist in the literature. In this paper, we propose a text line-level script identification technique written in *seven* popular official *Indic* scripts *namely*, *Gujarati*, *Kannada*, *Malayalam*, *Oriya*, *Tamil*, *Telugu*, *Urdu* along with *Roman* script.

## 2 Data Collection and Preprocessing

At present, no standard database of handwritten *Indic* scripts are available in public domain. Hence, we created our own database of handwritten documents in the laboratory. The document pages for the database are collected by different persons on request under our supervision. The writers are asked to write inside A-4 size pages, without imposing any constraint regarding the content of the textual materials. The document pages are digitized at 300 dpi resolution and stored as gray tone images. The scanned images may contain noisy pixels which are removed by applying Gaussian filter [11]. It should be noted that the handwritten text line (actually, portion of the line arbitrarily chosen) may contain two or more words with noticeable *intra-* and *inter-word* spacings. Numerals that may appear in the text are not considered for the present work. It is ensured that at least 50% of the cropped text line contains text. A sample snapshot of text line images written in eight different scripts is shown in Fig. 1. Otsu's global thresholding approach [12] is used to convert them into two-tone images. However, the dots and punctuation marks appearing in the text lines are not eliminated, since these may also contribute to the features of respective scripts. Finally, a total of 800 handwritten text line images are considered, with exactly100 text lines per script.

## 3 Feature Extraction

The feature extraction is based on the combination of Chain Code Histogram (CCH) and Discrete Fourier Transform (DFT) which are described in detail in the

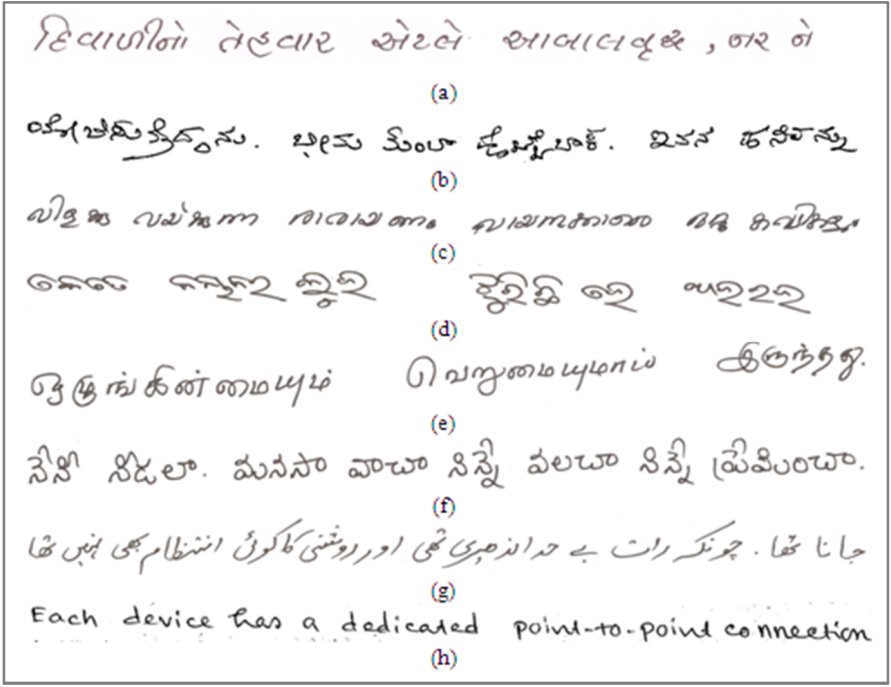

**Figure 1.** Sample text line images taken from our database written in: (a) *Gujarati*, (b) *Kannada*, (c) *Malayalam*, (d) *Oriya*, (e) *Tamil*, (f) *Telugu*, (g) *Urdu,* and (h) *Roman* scripts respectively

next subsection.

## 3.1 Chain Code Histogram

Chain codes [11] are used to represent a boundary by a connected sequence of straight-line segments of specified length and direction. It describes the movement along a digital curve or a sequence based on the connectivity. Two types of chain codes are possible which are based on the numbers of neighbors of a pixel, *namely*, four or eight, giving rise to 4- or 8-neighbourhood. The corresponding codes are the 4-directional code and 8-directional code, respectively. The direction of each segment is coded by using a numbering scheme as shown in Fig. 2. In the present work, the boundaries of handwritten text lines written in different scripts can be traced and allotted the respective numbers based on the directions. Thus, the boundary of each of the text line is reduced to a sequence of numbers. A boundary code formed as a sequence of such directional numbers is referred to as a *Freeman chain code*.

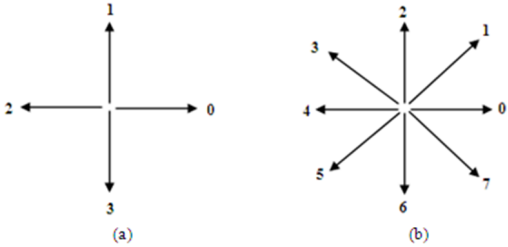

**Figure 2.** Illustration of numbering the directions for: (a) 4-dimensional, and (b) 8-dimensional chain codes

The histogram of *Freeman chain codes* are taken as feature values F1-F8 and the histogram of first difference of the chain codes are also taken as feature values F9-F15. Let us denote the set of pixels by R. The perimeter of a region R is the number of pixels present in the boundary of R. In a binary image, the perimeter is the number of foreground pixels that touches the background in the image. For an 8-directional code, the length of perimeter of each text line (F16) is calculated as: $|P|$ = Even count + $\sqrt{2}$ *(Odd count). A circularity measure (F17) proposed by Haralick [13] can be written as:

$$C_2 = \frac{\mu_R}{\sigma_R} \qquad (1)$$

where, $\mu_R$ and $\sigma_R$ are the mean and standard deviation of the distance from the centroid of the shape to the shape boundary and can be computed as follows:

$$\mu_R = \frac{1}{K} \sum_{k=0}^{K-1} \|(x_k, y_k) - (\bar{x}, \bar{y})\| \qquad (2)$$

$$\sigma_R = \left( \frac{1}{K} \sum_{k=0}^{K-1} [\|(x_k, y_k) - (\bar{x}, \bar{y})\| - \mu_R]^2 \right)^{1/2} \qquad (3)$$

where, the set of pixels $(x_k, y_k)$, $k = 0, \ldots, K-1$ lie on the perimeter P of the region. The circularity measure $C_2$ increases monotonically as the digital shape becomes more circular and is similar for digital and continuous shapes. Along the circularity, the slopes are labeled in accordance with their chain codes which are shown in Table 1.

**Table 1.** Labeling of slope angles according to their chain codes

| Chain code | 0 | 1 | 2 | 3 | 4 | 5 | 6 | 7 |
|---|---|---|---|---|---|---|---|---|
| θ | 0 | $45°$ | $90°$ | $135°$ | $180°$ | $-135°$ | $-90°$ | $-45°$ |

The count of the slopes having θ values $0°$, $|45°|$, $|90°|$, $|135°|$, $180°$ for each of the handwritten text line images are taken as feature values (F18-F22).

## 3.2 Discrete Fourier Transform

The Fourier Transform [11] is an important image processing tool which is used to decompose an image into its sine and cosine components. The output of the transformation represents the image in the Fourier or frequency domain, while the input image is the spatial domain equivalent. In the Fourier domain, each point in the spatial domain image represents a particular frequency.

The Discrete Fourier Transform (DFT) is the sampled Fourier Transform and therefore does not contain all frequencies forming an image, but only a set of samples which is large enough to fully describe the spatial domain image. The number of frequencies corresponds to the number of pixels in the spatial domain image, i.e., the images in the spatial and Fourier domains are of the same size. The DFT of a digital image of size $MxN$ can be written as:

$$G(u,v) = \frac{1}{MN} \sum_{m=0}^{M-1} \sum_{n=0}^{N-1} g(m,n) e^{-j2\pi\left(\frac{mu}{M} + \frac{nv}{N}\right)} \qquad (4)$$

where, $g(m,n)$ is the image in the spatial domain and the exponential term is the basis function corresponding to each point $G(u,v)$ in the Fourier space. The value of each point $G(u,v)$ is obtained by multiplying the spatial image with the corresponding base function and summing the result. The Fourier Transform produces a complex number valued output which can be displayed with two images, either with the real and imaginary parts or with the *magnitude* and *phase*, where *magnitude* determines the contribution of each component and *phase* determines which components are present. The plots for magnitude and phase components for a sample *Tamil* handwritten text-line image are shown in Fig. 3. In the current work, only the *magnitude* part of DFT is employed as it contains most of the information of the geometric structure of the spatial domain image. This in turn becomes easy to examine or process certain frequencies of the image. The *magnitude* coefficient is normalized as follows:

$$G'(u,v) = \frac{|G(u,v)|}{\sqrt{\sum_{u,v}|G(u,v)|^2}} \qquad (5)$$

The algorithm for feature extraction using DFT is as follows:

**Step 1:** Divide the input text line image into nxn non-overlapping blocks which are known as *grids*. The optimal value of $n$ has been chosen as 4.
**Step 2:** Compute the DFT (by applying Eqn. (4)) in each of the *grids*.
**Step 3:** Estimate only the *magnitude* part of the DFT and normalize it using Eqn. (5).
**Step 4:** Calculate the mean and standard deviation of the *magnitude* part from each of the *grids* which give a feature vector of 32 elements (F23-F54).

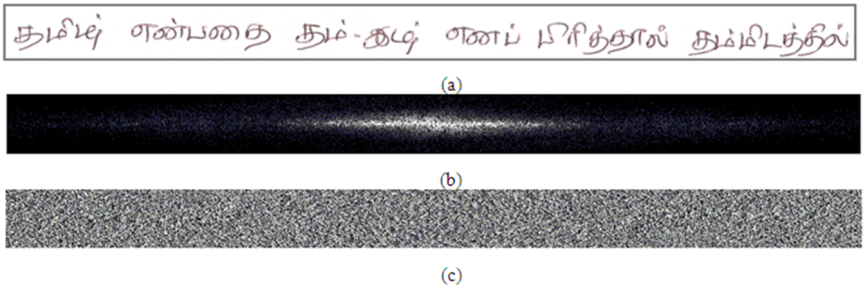

**Figure 3.** Illustration of: (a) handwritten *Tamil* text-line image, (b) its magnitude component, and (c) its phase component after applying DFT

## 4 Experimental Results and Discussion

The performance of the present script identification scheme is evaluated on a dataset of 800 preprocessed text line images as described in Section 2. For each dataset of 100 text line images of a particular script, 65 images are used for training and the remaining 45 images are used for testing purpose. The proposed approach is evaluated by using seven well-known classifiers *namely*, Naïve Bayes, Bayes Net, MLP, SVM, Random Forest, Bagging and MultiClass Classifier. The recognition performances and their corresponding scores achieved at 95% confidence level are shown in Table 2.

**Table 2.** Recognition performances of the proposed script identification technique using seven well-known classifiers (best case is shaded in grey and styled in bold)

|  | Classifiers | | | | | | |
| --- | --- | --- | --- | --- | --- | --- | --- |
|  | Naïve Bayes | Bayes Net | MLP | SVM | Random Forest | Bagging | MultiClass Classifier |
| Success Rate (%) | 89.33 | 90.09 | 95.14 | **97.03** | 94.6 | 91.25 | 92.74 |
| 95% confidence score (%) | 91.62 | 93.27 | 96.85 | **99.7** | 97.39 | 93.54 | 95.52 |

As observed from Table 2 that SVM classifier produces the highest identification accuracy of 97.03%. In the present work, detailed error analysis of SVM classifier with respect to different well-known parameters *namely*, Kappa statistics, mean absolute error, root mean square error, True Positive rate (TPR), False Positive rate (FPR), precision, recall, F-measure, Matthews Correlation Coefficient (MCC) and Area Under ROC (AUC) are also computed. The values of Kappa statistics, mean absolute error, root mean square error of SVM classifier for the present technique are found to be 0.9661, 0.0074 and 0.0862 respectively. Table 3 provides a statistical performance analysis of the remaining parameters for each of the aforementioned scripts.

**Table 3.** Statistical performance measures along with their respective means (shaded in grey and styled in bold) achieved by the proposed technique for eight handwritten scripts

| Scripts | TP rate | FP rate | Precision | Recall | F-measure | MCC | AUC |
| --- | --- | --- | --- | --- | --- | --- | --- |
| *Gujarati* | 1.000 | 0.000 | 1.000 | 1.000 | 1.000 | 1.000 | 1.000 |
| *Kannada* | 0.970 | 0.025 | 0.845 | 0.970 | 0.903 | 0.891 | 0.972 |
| *Malayalam* | 0.950 | 0.000 | 1.000 | 0.950 | 0.975 | 0.972 | 0.975 |
| *Oriya* | 1.000 | 0.000 | 1.000 | 1.000 | 1.000 | 1.000 | 1.000 |
| *Tamil* | 0.990 | 0.000 | 1.000 | 0.990 | 0.995 | 0.994 | 0.995 |
| *Telugu* | 0.980 | 0.000 | 1.000 | 0.980 | 0.990 | 0.989 | 0.990 |
| *Urdu* | 0.941 | 0.004 | 0.969 | 0.941 | 0.955 | 0.949 | 0.968 |
| *Roman* | 0.931 | 0.004 | 0.969 | 0.931 | 0.949 | 0.943 | 0.963 |
| **Weighted Average** | **0.970** | **0.004** | **0.973** | **0.970** | **0.971** | **0.967** | **0.983** |

Though Table 2 shows encouraging results but still some of the handwritten text lines are misclassified during the experimentation. The main reasons for the same are: (a) presence of speckled noise, (b) skewed words present in some text lines, and (c) occurrence of irregular spaces within text words, punctuation symbols, etc. The structural resemblance in the character set of some of the *Indic* scripts like *Kannada* and *Telugu* as well as *Malayalam* and *Tamil* causes similarity in the contiguous pixel distribution which in turns misclassifies them among each other. Fig. 4 shows some samples of misclassified text line images.

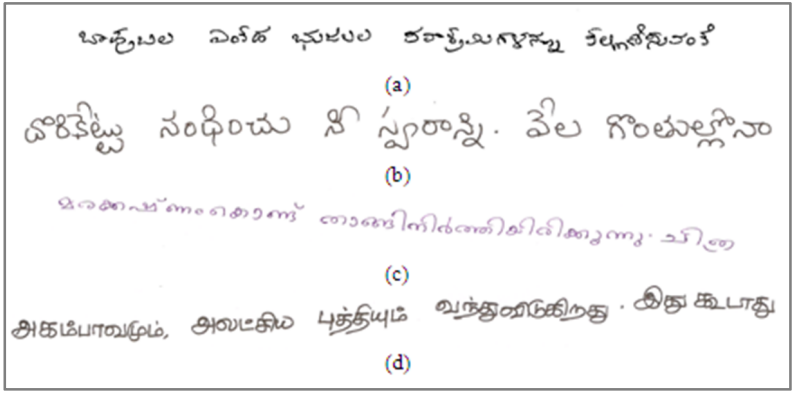

**Figure 4.** Samples of text line images written in (a) *Kannada*, (b) *Telugu*, (c) *Malayalam*, and (d) *Tamil* scripts misclassified as *Telugu*, *Kannada*, *Tamil* and *Malayalam* scripts respectively

## Conclusion

In this paper, we have proposed a robust method for handwritten script identification at text line-level for eight official scripts of India. The aim of this paper is to facilitate the research of multilingual handwritten OCR. A set of 54 feature values are extracted using the combination of CCH and DFT. Experimental results have shown that an accuracy rate of 97.03% is achieved using SVM classifier with limited dataset of eight different scripts which is quite acceptable taking the complexities and shape variations of the scripts under consideration. In our future endeavor, we plan to modify this technique to perform the script identification from handwritten document images containing more number of Indian languages. Another focus is to increase the size of the database to incorporate larger variations of writing styles which in turn would establish our technique as writer independent.

## Acknowledgment

The authors are thankful to the Center for Microprocessor Application for Training Education and Research (*CMATER*) and Project on Storage Retrieval and Understanding of Video for Multimedia (SRUVM) of Computer Science and Engineering Department, Jadavpur University, for providing infrastructure facilities during progress of the work. The current work, reported here, has been partially funded by University with Potential for Excellence (UPE), Phase-II, UGC, Government of India.